\documentclass{article}
\usepackage{ijcai26}

\usepackage{times}
\usepackage{url}
\usepackage[hidelinks]{hyperref}
\usepackage{iftex}
\ifPDFTeX
  \usepackage[utf8]{inputenc}
\fi
\usepackage[small]{caption}
\usepackage{graphicx}
\usepackage{amsmath}
\usepackage{amsthm}
\usepackage{booktabs}
\usepackage[table]{xcolor}
\usepackage{algorithm}
\usepackage{algorithmic}

\title{DiG-Plan: Mitigating Early Commitment for Tool-Graph Planning via \\ Diffusion Guidance}

\author{
Yansi Li
\and
Zhuosheng Zhang\thanks{Corresponding author. This work was supported by the National Key R\&D Program of China (No. 2024YFC3306500), the National Natural Science Foundation of China (62406188), and the Shanghai Municipal Special Program for Basic Research on General AI Foundation Models (2025SHZDZX025G08).}\\
\affiliations
School of Computer Science, Shanghai Jiao Tong University\\
\emails
yansi\_li@sjtu.edu.cn, zhangzs@sjtu.edu.cn
}

\begin{document}

\maketitle

\begin{abstract}
Generating executable tool plans requires selecting appropriate subsets from tool libraries, a combinatorial search problem with an exponentially large solution space. However, we identify a critical misalignment in predominant approaches: standard autoregressive (AR) decoding suffers from \textit{early commitment}, where initial token choices rigidly constrain the search trajectory. A controlled study shows that masked denoising raises Pass@10 solution coverage from 0.320 to 0.943 over AR sampling under matched compute. Motivated by this, we propose \textbf{DiG-Plan}, a framework that decouples combinatorial exploration from structural refinement. DiG-Plan employs a diffusion-based proposer to generate diverse tool sets via iterative refinement, followed by an AR refiner for dependency prediction. On TaskBench, DiG-Plan improves over AR baselines by a 10\% relative margin, with the largest gains on complex compositional tasks; API-Bank results show that the propose-refine-select design remains effective across domains. Code is available at \url{https://github.com/puddingyeah/DiG-Plan}.
\end{abstract}

\section{Introduction}
\label{sec:intro}

Tool-augmented large language models (LLMs) have evolved from solving simple queries to addressing complex, multi-step problems that require composing explicit executable plans~\cite{DBLP:conf/iclr/YaoZYDSN023,DBLP:conf/iclr/QinLYZYLLCTQZHT24}. Unlike single-tool invocation, real-world planning involves selecting an appropriate subset of tools from large libraries and determining their execution order~\cite{DBLP:conf/nips/0001STZRY00Z24}. This problem is fundamentally combinatorial: with a library of $N$ tools, the subset-hypothesis space contains $2^N$ possible tool sets. We do not assume exhaustive enumeration; the challenge is effective search in this large discrete space as libraries scale~\cite{DBLP:conf/nips/0001STZRY00Z24,DBLP:conf/emnlp/LiZ000YLHL23,DBLP:conf/iclr/0036YZXLL0DMYZ024}.

Despite the combinatorial nature of this task, predominant approaches treat tool planning as a sequential text generation problem, relying on standard autoregressive (AR) decoding~\cite{DBLP:conf/iclr/YaoZYDSN023,DBLP:conf/nips/YaoYZS00N23}. While AR models have proven highly effective for text generation, we identify a critical misalignment when applied to combinatorial search: \emph{early commitment}. Because AR generation proceeds strictly left-to-right, early token choices establish a prefix that constrains all subsequent decisions. Once the model commits to an initial tool selection, the probability distribution over remaining tools becomes conditioned on this prefix, limiting the model's ability to explore alternative tool combinations even under stochastic sampling.

This prefix constraint creates a form of \emph{path dependency} in the search process. Early choices determine the trajectory through the search space, effectively pruning large regions of the combinatorial space before they can be explored. For instance, we observe cases where AR models initially select generic vision tools for document-understanding tasks, failing to discover plans requiring more specialized document or visual question-answering tools even across multiple samples with varied sampling strategies. 

Crucially, we find that this limitation is resistant to simply increasing sampling diversity. Even with extensive stochastic sampling, AR models fail to recover the missed plans, indicating that the constraint is intrinsic to the sequential decoding mechanism itself rather than a lack of randomness.

We designed a synthetic tool-selection task to strictly disentangle the decoding mechanism from other variables, such as model architecture and pretraining data.
By fixing the output length and matching model capacity, we find that under identical compute budgets, masked denoising raises Pass@10 solution coverage from 0.320 to 0.943 compared with AR sampling. This provides empirical evidence that the limitation is intrinsic to the sequential decoding strategy itself.

Motivated by this insight, we propose \textbf{DiG-Plan} (\textbf{Di}ffusion-\textbf{G}uided \textbf{Plan}ning), a framework that addresses early commitment by \textit{decoupling} combinatorial exploration from structural refinement. DiG-Plan operates in three stages: first, a diffusion-based proposer generates diverse tool sets via iterative refinement, leveraging global context to revise early decisions; second, a shared autoregressive refiner predicts dependency structures conditioned on each proposed tool set, eliminating the combinatorial search problem in joint generation; finally, an inference-time judge-free value function selects the best candidate using only deployable features, without calling an external LLM judge.

\begin{figure*}[t]
\centering
\includegraphics[width=0.9\textwidth]{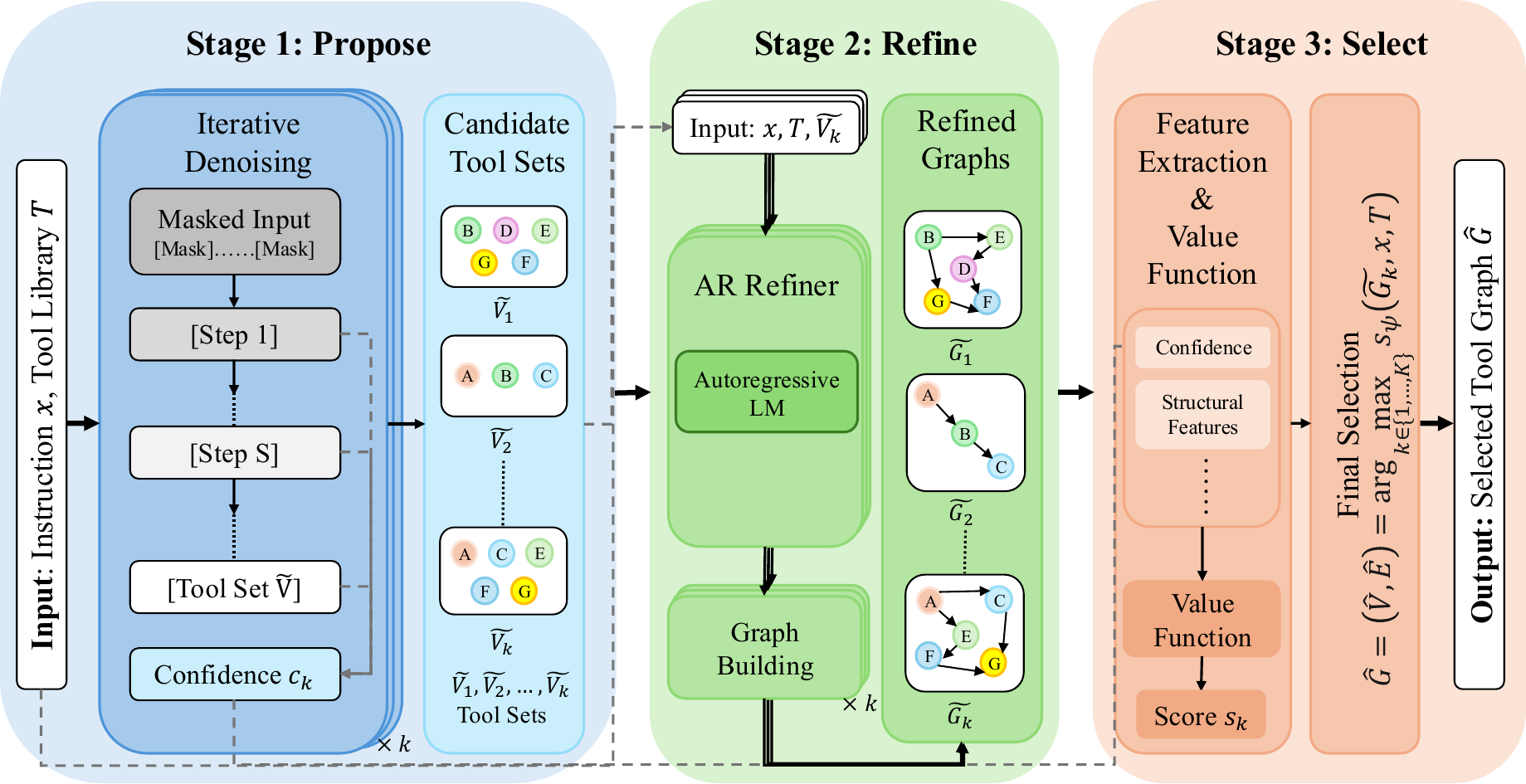}
\caption{DiG-Plan overview. (1) Propose: A diffusion-based model generates diverse candidate tool sets via iterative refinement; (2) Refine: A shared AR model predicts dependency edges for each fixed tool set; and (3) Select: An inference-time judge-free value function identifies the optimal plan using only deployable features, avoiding external LLM judging at deployment.}
\label{fig:digplan}
\end{figure*}

On TaskBench, DiG-Plan improves over AR baselines by a 10\% relative margin, with the largest gains on complex compositional tasks; API-Bank results show that the propose-refine-select design remains effective across domains. Diagnostic analyses show that diffusion proposers improve best-of-$K$ quality from 0.735 to 0.787 and union precision from 0.575 to 0.692 compared with AR proposers under matched compute budgets. Importantly, systematic sweeps over AR sampling parameters confirm that this advantage cannot be replicated by simply increasing sampling randomness, validating that the benefit stems from the iterative refinement mechanism rather than increased stochasticity.

Our main contributions are as follows:
\begin{itemize}
    \item We identify early commitment in autoregressive tool planning as a key barrier to combinatorial tool-set exploration, and validate this effect with a controlled fixed-length study showing that masked denoising covers substantially more valid tool-set hypotheses under matched compute.
    \item We introduce DiG-Plan, a propose-refine-select framework that separates tool-set exploration from dependency refinement: diffusion proposal searches over candidate tool subsets, autoregressive refinement predicts edges conditioned on each subset, and an inference-time judge-free value function selects among candidate graphs using deployable features.
    \item Through experiments on TaskBench, we show that DiG-Plan improves tool-set prediction over AR baselines by 10\%, while API-Bank results indicate cross-domain applicability. Candidate-pool diagnostics, AR sampling sweeps, and AR-beam comparisons confirm that the gains come from stronger proposal quality rather than simply increased stochasticity.
\end{itemize}

\section{Related Work}
\label{sec:related}

\subsection{Tool-Augmented LLMs}
Tool-augmented systems are typically formulated as reasoning-and-acting pipelines or tool-invocation learners~\cite{DBLP:conf/iclr/YaoZYDSN023,DBLP:conf/iclr/QinLYZYLLCTQZHT24}, often employing planner-centric orchestration for error correction~\cite{DBLP:journals/corr/abs-2511-10037,DBLP:conf/kdd/00060LHZC25}. Recent benchmarks like MetaTool and Toolink further diagnose tool choice behaviors~\cite{DBLP:conf/iclr/HuangSLFWZ000G024,DBLP:conf/naacl/QianXL024}. However, these methods predominantly rely on autoregressive decoding, which we show suffers from early commitment when the task requires exploring a combinatorial space of tool subsets.

\subsection{Planning Benchmarks}
Current benchmarks, including TaskBench, API-Bank, and AgentBench~\cite{DBLP:conf/nips/0001STZRY00Z24,DBLP:conf/emnlp/LiZ000YLHL23,DBLP:conf/iclr/0036YZXLL0DMYZ024}, have evolved to evaluate end-to-end planning capabilities. 
Since these metrics primarily prioritize final execution accuracy, the specific impact of decoding strategies on candidate diversity remains underexplored. We complement these benchmarks by focusing on their compositional subsets, conducting controlled studies to analyze how different decoding mechanisms influence exploration.

\subsection{Multi-Sample Reasoning}
Search-based prompting methods (e.g., Tree of Thoughts, Self-Consistency) and graph-structured reasoning~\cite{DBLP:conf/nips/YaoYZS00N23,DBLP:conf/iclr/0002WSLCNCZ23,DBLP:conf/aaai/BestaBKGPGGLNNH24,DBLP:journals/corr/abs-2502-05078} explore candidate sets to improve reasoning. 
These approaches effectively operate as search strategies atop autoregressive models. However, our experiments suggest that their performance is naturally bounded by the underlying proposal distribution. If prefix dependencies limit diversity, increasing the sampling budget offers limited gains. DiG-Plan addresses this limitation at the proposal stage by adopting diffusion models for non-autoregressive candidate generation.

\subsection{Diffusion and Denoising Language Models}
Diffusion models have introduced iterative refinement to text generation and discrete sequence modeling~\cite{DBLP:conf/nips/AustinJHTB21,DBLP:conf/nips/LiTGLH22,nie2025largelanguagediffusionmodels,DBLP:journals/corr/abs-2508-15487}. 
Most prior work has applied this capability to controllable generation or infilling tasks. Here, we extend it to combinatorial tool-set exploration. The global visibility of diffusion models facilitates non-sequential revisions, which is particularly helpful for handling complex inter-tool dependencies that are often challenging for strictly left-to-right generation.

\section{Problem Setup}
\label{sec:problem}

We now formalize the tool-graph planning problem and establish the evaluation framework.

\subsection{Tool-Graph Planning}
\label{sec:tool-graph}

A tool library $\mathcal{T}$ contains tools where each tool $t \in \mathcal{T}$ has a name and a short description. Given a natural-language instruction $x$, the goal is to predict a tool graph $\hat{G}=(\hat{V},\hat{E})$, where $\hat{V}\subseteq \mathcal{T}$ is the selected tool set and $\hat{E}\subseteq \hat{V}\times \hat{V}$ is a set of directed dependency edges. A directed edge $(u\!\rightarrow\!v)\in\hat{E}$ indicates that tool $v$ should be executed after tool $u$.

Evaluation compares predictions against a ground-truth tool graph $G^\star=(V^\star,E^\star)$ provided by the dataset~\cite{DBLP:conf/nips/0001STZRY00Z24}. Three factors make this problem challenging: (i) the tool library can be large, with a combinatorial search space of $2^{|\mathcal{T}|}$ possible subsets; (ii) instructions are often underspecified; and (iii) multiple valid plans may exist, but evaluation uses a single ground-truth reference.

\subsection{Metrics}
\label{sec:metrics}

\subsubsection{Tool-Set F1}
We denote this metric as ToolF1. Given predicted tool set $\hat{V}$ (canonicalized against $\mathcal{T}$) and ground-truth $V^\star$:
\begin{align}
  P_V &= \frac{|\hat{V}\cap V^\star|}{\max(|\hat{V}|,1)}, \quad
  R_V = \frac{|\hat{V}\cap V^\star|}{\max(|V^\star|,1)}, \notag \\
  \mathrm{ToolF1}(\hat{V},V^\star) &= \frac{2P_VR_V}{\max(P_V+R_V,\epsilon)},
\end{align}
where $\epsilon=10^{-8}$ ensures numerical stability.

\subsubsection{Edge Recall}
We denote this metric as EdgeRec. Given predicted edge set $\hat{E}$ (restricted to endpoints in $\hat{V}$) and ground-truth $E^\star$:
when $|E^\star|=0$, EdgeRec is 1 if $|\hat{E}|=0$ and 0 otherwise; when $|E^\star|>0$,
\begin{equation}
  \mathrm{EdgeRec}(\hat{E},E^\star)=\frac{|\hat{E}\cap E^\star|}{|E^\star|}.
\end{equation}

\subsubsection{Candidate-Pool Metrics}

To diagnose candidate diversity and upper-bound quality, we use three complementary metrics. Given a pool of $K$ candidates $\{(\tilde{V}_k,\tilde{E}_k)\}_{k=1}^K$:
\begin{align}
  \mathrm{Pass@K} &= \mathrm{UnionRecall@K}
  =\frac{\left|\left(\cup_{k=1}^K \tilde{V}_k\right)\cap V^\star\right|}{\max(|V^\star|,1)}, \\
  \mathrm{UnionPrec@K} &=
  \frac{\left|\left(\cup_{k=1}^K \tilde{V}_k\right)\cap V^\star\right|}{\max\left(\left|\cup_{k=1}^K \tilde{V}_k\right|,1\right)}, \\
  \mathrm{Oracle@K} &= \max_{1\le k\le K}\ \mathrm{ToolF1}(\tilde{V}_k,V^\star).
\end{align}
Pass@K quantifies the aggregate \emph{coverage} of the generated pool by computing the union recall against the ground truth. A high value confirms that the model has effectively explored the relevant search space. Oracle@K establishes the performance \emph{upper bound} (best-of-$K$) by selecting the optimal candidate via ground-truth labels. The difference between these two metrics isolates the error source: low Pass@K implies poor exploration, while a large gap to the \emph{upper bound} indicates poor selection.

\subsubsection{Executability}
We define a lightweight predicate, $\mathrm{ExecOK}(\hat{G})\in\{0,1\}$, to enforce structural validity using the predicted graph.
It returns 1 only if three conditions are met: (i) all edge endpoints lie in the predicted node set, with no self-loops or duplicate edges; (ii) for chain instances, edges form a single path visiting all nodes; and (iii) for DAG tasks, the graph contains at least one sink and no isolated nodes. We apply this strictly as an optional filter during selection to ensure deployability.

\subsection{Motivating Evidence: A Controlled Study}
\label{sec:controlled}

To validate the early commitment hypothesis, we construct a controlled tool-set prediction task that isolates the combinatorial search component. We define a tool universe $\mathcal{T}=\{t_1,\ldots,t_{23}\}$ and represent tool subsets as fixed-length 23-bit vectors. Each sample provides a symbolic graph specification (sources, sinks, and a noisy depth hint), and the model predicts which tools are needed.
This task has the same combinatorial structure: identifying a sparse subset from $2^{23}$ possible combinations.

We control three key factors: (i) \emph{model capacity} by using identical small Transformer backbones with 2 layers, 128 hidden dimensions, and 4 attention heads for both AR and denoising models; (ii) \emph{training data} by generating a shared synthetic corpus from the same distribution; and (iii) \emph{output format} by fixing the output length to 23 bits, removing any advantage from variable-length generation. The only manipulated factor is the decoding family: prefix AR through next-token prediction versus masked denoising, which randomly masks 30\% of bits and predicts them conditioned on unmasked context.

We generate $K{=}10$ candidates per input using temperature sampling for AR and stochastic denoising for masked models, then evaluate Pass@K and Oracle@K.

\begin{table}[t]
\centering
\small
\begin{tabular}{lccc}
\toprule
Model & k=1 ToolF1 & Pass@10 & Oracle@10 \\
\midrule
Greedy AR & 0.355$\pm$0.00 & 0.320$\pm$0.00 & 0.355$\pm$0.00 \\
Masked denoising & 0.349$\pm$0.04 & 0.943$\pm$0.02 & 0.566$\pm$0.01 \\
\bottomrule
\end{tabular}
\caption{Controlled tool-set playground with 23-bit outputs and $K{=}10$ candidates. Values are mean$\pm$std over 6 seeds.}
\label{tab:playground}
\end{table}

Table~\ref{tab:playground} shows that masked denoising raises Pass@10 coverage from 0.320 to 0.943 and Oracle@10 from 0.355 to 0.566, while single-candidate quality remains comparable. With 10 candidates, masked denoising covers 94\% of the ground-truth tools on average, while AR covers only 32\%. Since model capacity, training data, and output format are controlled, this result provides direct evidence that iterative denoising explores a broader region of the tool-set space than prefix AR decoding.

\paragraph{Implications.} This controlled study provides empirical evidence that AR's sequential generation limits exploration in combinatorial search spaces. Motivated by this finding, we propose DiG-Plan (Section~\ref{sec:method}), which leverages diffusion language models for diverse candidate proposal while using AR for structural refinement.

\section{Method: DiG-Plan}
\label{sec:method}

\subsection{Probabilistic Decomposition}
\label{sec:decomposition}

The controlled study in Section~\ref{sec:controlled} demonstrates that diffusion-based decoding achieves higher exploration diversity than AR in combinatorial search spaces. 
Motivated by this finding, DiG-Plan follows a propose-refine-select paradigm that separates tool selection from dependency prediction.
Given an instruction $x$ and a tool library $\mathcal{T}$, the framework operates in three stages (Figure~\ref{fig:digplan}). First, generate $K$ diverse candidate tool sets $\{\tilde{V}_k\}_{k=1}^K$ through diffusion-based proposal. Second, refine each candidate into a directed tool graph $\tilde{G}_k=(\tilde{V}_k,\tilde{E}_k)$ by predicting dependency edges with autoregressive decoding. Third, select a single plan $\hat{G}$ using a deployable value function.

DiG-Plan splits tool-graph planning into two stages:
\begin{align}
  \text{(Tool Set)}\quad &\tilde{V} \sim p_\phi(V \mid x,\mathcal{T}), \\
  \text{(Dependency)}\quad &\tilde{E} \sim p_\theta(E \mid x,\mathcal{T},\tilde{V}),
\end{align}
where $\phi$ denotes the proposer parameters and $\theta$ denotes the refiner parameters. 
The first stage explores diverse tool combinations using diffusion language models. The second stage predicts dependencies using autoregressive decoding, which naturally captures sequential structure.
The first stage focuses purely on exploring diverse tool combinations, while the second stage focuses on structured prediction given a fixed tool set. 
This separation allows using diffusion language models for the first stage and autoregressive decoding for the second stage.

\subsection{Diffusion Proposer}
\label{sec:proposer}

As established in Section~\ref{sec:intro}, autoregressive decoding suffers from early commitment in combinatorial search spaces. Diffusion language models~\cite{DBLP:conf/nips/AustinJHTB21,DBLP:conf/nips/LiTGLH22,nie2025largelanguagediffusionmodels,DBLP:journals/corr/abs-2508-15487} offer an alternative: they iteratively refine a complete output through multiple denoising steps, allowing the model to revise early decisions.

The proposer defines a distribution over tool subsets, instantiated with a diffusion language model. Stochastic sampling draws $K$ candidates $\{\tilde{V}_1,\ldots,\tilde{V}_K\}$ under fixed compute budget.
Our controlled experiments (Section~\ref{sec:controlled}) show that diffusion models maintain higher hypothesis diversity than AR decoding under matched budgets. We compare against AR proposers with temperature and nucleus sampling sweeps in Section~\ref{sec:experiments}. Even with top-$p{=}1.0$, AR does not match diffusion's exploration quality.

For diffusion proposers exposing per-step token entropies, we derive plan-level confidence by aggregating entropy over the diffusion trajectory. Let $H^{(s)}$ denote mean entropy over masked positions at step $s$. 
We discard the first 3 burn-in steps and define $\bar{H}=\frac{1}{S-S_0}\sum_{s=S_0+1}^{S} H^{(s)}$ and $c=\frac{1}{1+\bar{H}}\in(0,1]$ as input to the value function, where $S_0=3$.

\subsection{Autoregressive Refiner}
\label{sec:refiner}

While diffusion language models excel at exploring discrete combinatorial spaces, autoregressive decoding naturally captures sequential dependencies through its left-to-right generation process. By conditioning on a fixed tool set $\tilde{V}_k$, we eliminate the combinatorial search problem that causes early commitment.

For each candidate $\tilde{V}_k$, an autoregressive refiner predicts dependency structure $\tilde{E}_k$ conditioned on $(x,\mathcal{T},\tilde{V}_k)$. The refiner is shared across all proposers to isolate tool-set exploration effects. It takes as input the instruction $x$, the tool library $\mathcal{T}$, and the proposed tool set $\tilde{V}_k$, then generates a structured output specifying dependencies between tools. The output is parsed into tool graph $\tilde{G}_k=(\tilde{V}_k,\tilde{E}_k)$ after canonicalizing tool names and filtering edges.

\begin{table*}[t]
\centering
\small
\setlength{\tabcolsep}{12pt}
\begin{tabular}{llcc}
\toprule
Proposer & Refiner & ToolF1 $\uparrow$ & EdgeRec $\uparrow$ \\
\midrule
\multicolumn{4}{l}{\emph{Autoregressive and retrieval baselines}} \\
\addlinespace[0.25em]
AR-only & -- & 0.629$\pm$0.30 & 0.200$\pm$0.36 \\
AR proposer & AR refiner & 0.661$\pm$0.32 & 0.242$\pm$0.38 \\
Retriever (top-$k{=}5$) & AR refiner & 0.569 & 0.127 \\
Retriever (top-$k{=}10$) & AR refiner & 0.638 & 0.156 \\
Retriever (top-$k{=}15$) & AR refiner & 0.660 & 0.185 \\
\midrule
\multicolumn{4}{l}{\emph{Diffusion proposers}} \\
\addlinespace[0.25em]
Dream DLM-only & -- & 0.128$\pm$0.28 & 0.335$\pm$0.47 \\
LLaDA proposer & AR refiner & 0.734 & 0.258 \\
\textbf{Dream proposer} & \textbf{AR refiner} & \textbf{0.729$\pm$0.28} & \textbf{0.268$\pm$0.39} \\
\bottomrule
\end{tabular}
\caption{End-to-end tool-graph planning on TaskBench-23 with $N{=}501$ and $K{=}1$. We separate the \emph{proposer}, the tool-set generator, from the \emph{refiner}, the edge predictor. Numbers are means over instances, with standard deviations shown when available.}
\label{tab:taskbench-main}
\end{table*}

\subsection{Judge-Free Selection via a Value Function}
\label{sec:selection}

Best-of-$K$ gains require a selector to choose among candidates. 
Oracle selection is not deployable, as it requires access to ground-truth labels at inference time. External LLM judges introduce both cost and evaluation confounds. We train a lightweight value function $s_\psi(\tilde{G},x,\mathcal{T})$ that scores candidate graphs using only deployable, candidate-observable features.

At inference time, we select the highest-scoring candidate:
\begin{equation}
  \hat{G} = \arg\max_{k\in\{1,\ldots,K\}} s_\psi(\tilde{G}_k,x,\mathcal{T}).
\end{equation}

\paragraph{Training objective.} On a training split where ground-truth graphs are available, the value function is trained to predict a combined target that trades off tool-set accuracy and edge coverage:
\begin{equation}
  y(\tilde{G},G^\star)=\alpha\cdot \mathrm{ToolF1}(\tilde{V},V^\star) + (1-\alpha)\cdot \mathrm{EdgeRec}(\tilde{E},E^\star),
\end{equation}
with $\alpha=0.7$ by default. This weighting prioritizes tool-set accuracy: correct tool selection is required for accurate edge prediction. The predictor $s_\psi$ is fit by regression to $y$ using only candidate-observable features.

\paragraph{Feature design.} The input to $s_\psi$ concatenates several feature groups:
\begin{itemize}
    \item \emph{Proposer confidence} $c$, derived from diffusion entropy to capture uncertainty.
    \item \emph{Structural heuristic score} combining three components with weights 0.3/0.3/0.4: structural validity, coverage, and coherence.
    \item \emph{Predicted counts} (number of tools and edges).
    \item \emph{DAG-validity flag} computed by topological sorting.
    \item \emph{Candidate index} $k$ to capture position bias.
    \item \emph{Multi-hot tool vector} over predicted tool names to capture tool-specific patterns.
\end{itemize}
In the \texttt{graph} feature set, we additionally append a directed edge multi-hot vector over tool pairs. Importantly, no ground-truth-derived quantities are used at inference time, so the selector is fully deployable without oracle access.

\subsection{Executability Constraints}
\label{sec:exec}

When high-scoring candidates are close, exec-filtered selection chooses the highest-scoring candidate satisfying $\mathrm{ExecOK}(\tilde{G}_k){=}1$ (as defined in Section~\ref{sec:problem}), falling back to unconstrained top-1 otherwise.

\section{Experiments}
\label{sec:experiments}

We evaluate on TaskBench-23~\cite{DBLP:conf/nips/0001STZRY00Z24} and API-Bank~\cite{DBLP:conf/emnlp/LiZ000YLHL23}, reporting end-to-end tool-graph metrics (ToolF1 and EdgeRec) and candidate-pool diagnostics (Pass@K, Oracle@K).

\begin{table*}[t!]
\centering
\small
\setlength{\tabcolsep}{5pt}
\begin{tabular}{lcccc}
\toprule
Proposer family & $k{=}1$ ToolF1 & Oracle@10 & UnionRec@10 & UnionPrec@10 \\
\midrule
AR proposer ($T{=}1.3$) & 0.603 & 0.735 & 0.728 & 0.575 \\
Dream proposer & 0.683 & 0.787 & 0.759 & 0.692 \\
LLaDA2 proposer & 0.535 & 0.605 & 0.547 & 0.612 \\
\bottomrule
\end{tabular}
\caption{Candidate-pool diagnostics on the held-out chain+DAG set ($N{=}334$) with $K{=}10$ candidates per input.}
\label{tab:pool-diagnostics}
\end{table*}

\begin{table}[t!]
\centering
\small
\setlength{\tabcolsep}{4pt}
\begin{tabular}{lcc}
\toprule
Selector & ToolF1 $\uparrow$ & EdgeRec $\uparrow$ \\
\midrule
$K{=}1$   & 0.685 & 0.244 \\
Heuristic & 0.690 & 0.311 \\
\textbf{Value function} & \textbf{0.716} & \textbf{0.309} \\
Oracle    & 0.768 & 0.283 \\
\bottomrule
\end{tabular}
\caption{Selection on the held-out compositional subset ($N{=}334$) with $K{=}5$ candidates.}
\label{tab:selection}
\end{table}

\begin{figure*}[t]
\centering
\begin{minipage}{0.48\textwidth}
    \centering
    \includegraphics[width=0.9\linewidth]{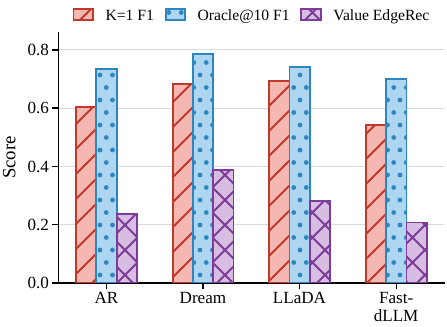}
    \vspace{-8pt}
    \caption*{\small (a) TaskBench-23 proposer comparison}
\end{minipage}
\hfill
\begin{minipage}{0.48\textwidth}
    \centering
    \includegraphics[width=0.9\linewidth]{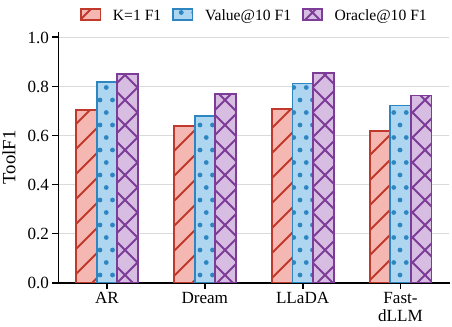}
    \vspace{-8pt}
    \caption*{\small (b) API-Bank cross-domain results}
\end{minipage}
\vspace{-5pt}
\caption{Proposer family comparison. (a) On TaskBench-23 compositional instances with $N{=}334$ and $K{=}10$, Dream improves over AR on $K{=}1$ ToolF1, Oracle@10, and EdgeRec. (b) On API-Bank, LLaDA reaches the strongest Oracle@10 and competitive value-selected ToolF1, averaged over 4 seeds.}
\label{fig:proposer-comparison}
\end{figure*}

\subsection{Benchmarks}
\label{sec:benchmarks}

\paragraph{TaskBench-23.} TaskBench~\cite{DBLP:conf/nips/0001STZRY00Z24} provides a tool-graph planning benchmark with diverse dependency structures. We use $N{=}501$ instances, with 167 each for single-tool, chain, and DAG tasks. For candidate-pool analyses, we focus on the compositional subset of chain and DAG instances, yielding $N{=}334$ instances.

\paragraph{API-Bank.} We convert API-Bank~\cite{DBLP:conf/emnlp/LiZ000YLHL23} to TaskBench format with tool nodes and dependency edges, serving as an out-of-domain check.

\subsection{Baselines}
\label{sec:baselines}

We compare DiG-Plan against several baselines: \textbf{AR-only}, \textbf{AR two-stage}, \textbf{AR-beam+AR}, \textbf{Retriever+AR} (dense retriever with $k\in\{5,10,15\}$ combined with AR refiner), and \textbf{Diffusion+AR}. The retriever baseline uses a pre-trained dense retriever (all-MiniLM-L6-v2) without task-specific finetuning, computing cosine similarity between instruction embeddings and tool description embeddings to select top-k tools. All two-stage methods share the same AR refiner to isolate the proposer's effect.

\subsection{Implementation Details}
\label{sec:impl}

\paragraph{Model selection.} All two-stage methods share the same AR refiner, differing only in the proposer. For end-to-end evaluation we report $K{=}1$ in Table~\ref{tab:taskbench-main}, while for candidate-pool analyses we generate $K\in\{5,10\}$ candidates per input. The value function is trained on a disjoint training split with supervision $y=\alpha\cdot\mathrm{ToolF1}+(1-\alpha)\cdot\mathrm{EdgeRec}$ where $\alpha=0.7$.

\paragraph{Proposer models.} Diffusion proposers use Dream 7B~\cite{DBLP:journals/corr/abs-2508-15487}, LLaDA-8B-Instruct~\cite{nie2025largelanguagediffusionmodels}, or LLaDA2.0-mini-preview~\cite{bie2025llada20scalingdiffusionlanguage}. AR components use Qwen2.5-7B-Instruct~\cite{DBLP:journals/corr/abs-2412-15115}. The retriever baseline uses all-MiniLM-L6-v2~\cite{DBLP:conf/emnlp/ReimersG19}.

\paragraph{Value function.} The value function uses a GradientBoostingRegressor~\cite{DBLP:journals/jmlr/PedregosaVGMTGBPWDVPCBPD11} with 300 trees, learning rate 0.05, and max depth 4, trained on candidate pools from the training split.

Table~\ref{tab:pool-diagnostics} presents candidate-pool diagnostics on the held-out chain+DAG subset with $N{=}334$ and $K{=}10$. Compared with an AR proposer under comparable budgets, Dream raises UnionPrec@10 from 0.575 to 0.692 and Oracle@10 from 0.735 to 0.787, indicating both stronger best-of-$K$ quality and higher-precision coverage.

\begin{figure*}[t]
\centering
\includegraphics[width=\textwidth]{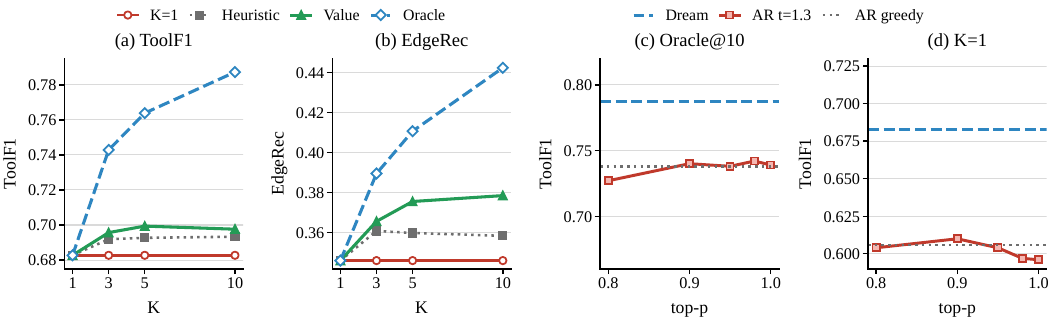}

\caption{Selection and sampling analysis. (a--b) Performance on the held-out compositional subset with $N{=}334$ under a fixed Dream-proposed pool as $K$ varies. (c--d) AR top-$p$ sweep on TaskBench-23 compositional instances with $N{=}334$ and $K{=}10$.}
\label{fig:selection-and-sweep}
\end{figure*}

\begin{table*}[tb!]
\centering
\footnotesize
\setlength{\tabcolsep}{2pt}
\newcommand{\cor}[1]{\textcolor{green!60!black}{#1}}
\newcommand{\err}[1]{\textcolor{red!80!black}{#1}}
\begin{tabular}{@{}>{\raggedright\arraybackslash}p{0.055\textwidth} >{\raggedright\arraybackslash}p{0.415\textwidth} >{\raggedright\arraybackslash}p{0.245\textwidth} >{\raggedright\arraybackslash}p{0.245\textwidth}@{}}
\toprule
\textbf{Type} & \textbf{Task Setup (Instruction \& Ground Truth)} & \textbf{Dream (Ours)} & \textbf{AR (Baseline)} \\
\midrule
\textbf{Chain} & 
\textit{I have a Spanish quote... translate it, then generate creative text, and finally analyze the document image.} \par
\vspace{3pt}
\textbf{GT:} Translation, Text Generation, Document QA & 
\cor{Translation, Text Generation, Document QA} \newline \textbf{(F1: 1.00)} & 
\cor{Translation}, \err{Conversational}, \err{Image Classification} \newline \textit{(F1: 0.33)} \\
\midrule
\textbf{DAG} & 
\textit{I have a long article... summarize it, generate an image, and check whether the image includes a playground.} \par
\vspace{3pt}
\textbf{GT:} Summarization, Text-to-Image, Visual QA & 
\cor{Summarization, Text-to-Image, Visual QA} \newline \textbf{(F1: 1.00)} & 
\err{Image-to-Image}, \cor{Summarization}, \err{Document QA} \newline \textit{(F1: 0.33)} \\
\bottomrule
\end{tabular}
\caption{Representative case studies comparing Dream and AR. \textcolor{green!60!black}{Green text} denotes correctly predicted steps matching the Ground Truth (GT), while \textcolor{red!80!black}{red text} indicates selection errors.}
\label{tab:case-study}
\end{table*}
\subsection{Disentangling Proposal Quality from Selection}
\label{sec:selection-results}

Table~\ref{tab:taskbench-main} reports end-to-end results on TaskBench-23 with $N{=}501$ and $K{=}1$. 
DiG-Plan with Dream proposer achieves ToolF1 of 0.729 and EdgeRec of 0.268, outperforming the AR two-stage baseline with ToolF1 of 0.661 and EdgeRec of 0.242, corresponding to a 10\% relative improvement. 
Since single-tool instances do not require combinatorial search, performance differences mainly arise on chain and DAG instances; we therefore focus additional analyses on the compositional subset of $N{=}334$ instances.

Several trends emerge from these results. First, two-stage methods outperform single-stage AR-only baseline, which achieves ToolF1 of 0.629. Second, among two-stage methods, diffusion proposers outperform AR proposers and retrieval-based proposers. Third, the Dream DLM-only baseline achieves high EdgeRec of 0.335 but low ToolF1 of 0.128. However, this aggregate EdgeRec is driven by single-tool instances: DLM-only achieves near-perfect EdgeRec of 0.994 on single tasks but fails on chain with 0.006 and DAG with 0.004 structures. This shows that end-to-end diffusion generation struggles with precise tool-name grounding and complex dependency prediction on compositional tasks, motivating our two-stage design where diffusion handles tool-set exploration and AR handles edge prediction.

Figure~\ref{fig:proposer-comparison} visualizes this advantage across multiple metrics. On the compositional subset (Figure~\ref{fig:proposer-comparison}a), Dream improves over AR on $K{=}1$ ToolF1, Oracle@10, and EdgeRec, matching the candidate-pool gains in Table~\ref{tab:pool-diagnostics}. Figure~\ref{fig:proposer-comparison}b further shows that the propose-refine-select interface transfers to API-Bank, where LLaDA achieves the strongest Oracle@10 and competitive value-selected ToolF1.

We further test whether the gap is due to weak AR decoding by evaluating an AR-beam proposer under the same held-out protocol, using $N{=}334$, $K{=}10$, and the same execution-constrained selector. AR-beam reaches ToolF1/EdgeRec of 0.625/0.294, still below Dream's 0.708/0.398, showing that stronger AR search alone does not close the gap.

Having established the effectiveness of the diffusion proposer, we next investigate the impact of the selection mechanism. To disentangle the contributions of proposal quality from selection strategy, we fix a single Dream-proposed candidate pool with the shared AR refiner at $K=5$ and systematically vary only the selection strategy on the held-out compositional test set of $N=334$ instances. This experimental design isolates the selector's contribution, allowing us to measure selection quality independently of proposal quality.

Table~\ref{tab:selection} demonstrates that an inference-time judge-free value function achieves ToolF1 of 0.716 compared to 0.690 for heuristic selection. Measured relative to the $K{=}1$ baseline of 0.685, this recovers $0.031/0.083\approx 37\%$ of the oracle gap to the best-of-5 value of 0.768.

Figure~\ref{fig:selection-and-sweep} shows (a) best-of-$K$ scaling and (b) AR sampling sweep results. The value function consistently outperforms heuristic selection across all $K$ values, and AR cannot match Dream's performance even when higher top-$p$ values increase sampling diversity.

\subsection{Error Analysis and Complexity Stratification}
We analyze failure patterns using an automatic multi-label taxonomy on the held-out compositional subset of $N=334$ instances. The taxonomy evaluates the value-function-selected predictions from a Dream proposer pool with $K=10$ candidates. Error tags are not mutually exclusive, so each predicted graph can receive multiple labels.

Missing edges are the most frequent error tag, affecting 82.9\% of samples; missing tools follow at 76.9\%. This indicates that coverage remains the primary challenge. Better tool proposals are still useful: Dream raises exact-tool coverage from 0.234 to 0.296 and exact+full-edge coverage from 0.141 to 0.231 over AR. Since the refiner cannot recover edges between missing tools, proposal and refinement quality are complementary. With ground-truth tools fixed, an oracle-refiner diagnostic raises EdgeRec from 0.398 to 0.675, identifying edge refinement as the remaining bottleneck. Extra tools appear in 50.9\% of predictions, and overt cyclicity remains rare at 6.9\%.

Using $K=5$ selection from the same Dream pool, value-function selection typically provides gains over heuristic selection, with notable improvements on instances with 3-4 tools where combinatorial search is non-trivial. For example, on chain instances with 3-4 tools, value-function selection achieves ToolF1 of 0.680 and 0.636 respectively, compared to heuristic's 0.661 and 0.632.

\subsection{Qualitative Analysis}
\label{sec:qualitative}

Table~\ref{tab:case-study} gives representative chain and DAG examples under the same refiner and value-based selector. Dream identifies all required tools in both cases, while AR makes semantically plausible but task-mismatched substitutions such as \textit{Image Classification} for \textit{Document QA} and \textit{Image-to-Image} for \textit{Text-to-Image}. These examples illustrate how early prefix choices can steer AR away from better tool combinations.

\section{Conclusion}
\label{sec:conclusion}

We presented DiG-Plan, a diffusion-guided propose-refine-select framework for tool-graph planning. The central finding is that prefix autoregressive decoding suffers from early commitment in tool-set search: once early tool choices are sampled, subsequent decisions remain conditioned on that prefix, limiting the diversity of candidate tool sets. DiG-Plan addresses this mismatch by using diffusion for global tool-set proposal, AR decoding for dependency refinement, and an inference-time judge-free value function for candidate selection.

Controlled and benchmark evaluations show stronger candidate coverage and a 10\% ToolF1 gain over AR baselines. Candidate-pool diagnostics, AR sweeps, and AR-beam comparisons attribute these gains to proposal quality rather than stochasticity. Future work will strengthen edge refinement.

\bibliographystyle{named}
\bibliography{references}

\end{document}